%% file: iclr2026_conference.tex
\documentclass{article} 
\usepackage{iclr2026_conference,times,enumitem}

\input{math_commands.tex}

\usepackage{hyperref}
\usepackage{url}
\usepackage{graphicx}
\usepackage{listings}
\usepackage{booktabs}
\usepackage{multirow}

\newcommand{\equalcontrib}{\thanks{Equal contribution.}}

\title{Evaluating Ill-Defined Tasks in Large Language Models}

\author{Yi Zhou\equalcontrib\\
IBM Research \\
San Jose, CA \\
\texttt{yi.zhou@ibm.com} \\
\And
Basel Shbita\footnotemark[1]\\
IBM Research \\
San Jose, CA \\
\texttt{basel@ibm.com} \\
}

\iclrfinalcopy 
\begin{document}

\maketitle

\input{latex/abstract}
\input{latex/intro}
\input{latex/ill_defined}
\input{latex/case_studies}
\input{latex/discussion}

\bibliography{iclr2026_conference}
\bibliographystyle{iclr2026_conference}

\appendix
\input{latex/appendix}

\end{document}

%% file: math_commands.tex
\usepackage{amsmath,amsfonts,bm}

\def\eqref#1{equation~\ref{#1}}

\def\1{\bm{1}}

\DeclareMathAlphabet{\mathsfit}{\encodingdefault}{\sfdefault}{m}{sl}
\SetMathAlphabet{\mathsfit}{bold}{\encodingdefault}{\sfdefault}{bx}{n}



%% file: latex/abstract.tex
\begin{abstract}
Many evaluations of Large Language Models (LLMs) target tasks that are inherently ill-defined, with unclear input and output spaces and ambiguous success criteria.
We analyze why existing evaluation benchmarks and metrics fail to provide reliable or diagnostic signals of model capability for such tasks.
We examine two case studies: Complex Instruction Following (CIF), where we identify recurring issues including limited coverage of real-world instruction complexity, sensitivity to instruction phrasing, inconsistent and non-comparable metrics, and instability introduced by LLM-based judges; and Natural Language to Mermaid Sequence Diagrams (NL2Mermaid), where we show how multi-faceted evaluation criteria can yield actionable insights beyond aggregate scores.
Together, these case studies show that current evaluations frequently conflate distinct failure modes, yielding scores that are unstable, non-diagnostic, and difficult to act upon.
Our findings expose fundamental limitations in existing evaluation practices for ill-defined tasks and motivate more robust, interpretable evaluation designs.
\end{abstract}

%% file: latex/intro.tex
\section{Introduction}
\label{sec:intro}

Large language models (LLMs) have made rapid progress in artificial intelligence, driving significant advances in natural language understanding and reasoning, and demonstrating strong performance across a broad range of tasks~\citep{naveed2025comprehensive,matarazzo2025survey, yang2024harnessing,brown2020language,wang2019superglue}.
They now support a wide range of workflows, spanning domain-specific fine-tuning pipelines~\citep{ouyang2022training} to agent-based systems capable of multi-step planning and tool use~\citep{wu2025agentic}.
Despite their success, LLMs pose unique challenges due to probabilistic outputs, opaque behavior, and hard-to-isolate failure modes~\citep{gao2025llm}.
As their use grows, evaluation benchmarks must extend beyond performance metrics to diagnose failures and suggest directions for improvement.

In this paper, we focus on a class of problems known as ill-defined tasks, in which goals, constraints, or evaluation criteria are under-specified or open to interpretation.
Such tasks are pervasive in real-world applications, e.g., the case of Complex Instruction Following (CIF), where multiple solutions may be acceptable and the boundaries between success and failure are often very subjective. 
By studying LLM performance on ill-defined tasks, we can gain deeper insights into how they handle user instructions, why models fail, and what signals can guide the development of more robust and interpretable systems.
However, most existing benchmarks are designed for well-defined tasks with a unique or narrowly specified notion of correctness~\citep{hendryckstest2021,hendrycks2021ethics,wang2024mmlu}.
Benchmarks for ill-defined tasks fail to account for ambiguity and the potentially infinite number of valid solutions, and provide limited visibility into intermediate reasoning or failure modes~\citep{chang2024survey,li2024llms,wang2025can}.
To address this gap, we analyze why existing evaluation approaches for ill-defined tasks fail to provide reliable or diagnostic signals of model capability.
We focus on three primary failure sources: (i) benchmark data lacking real-world coverage and constraint diversity, (ii) metrics that collapse multiple diagnostic dimensions into single scores, conflating failure modes, and (iii) variance and bias introduced by LLM-as-a-Judge (LLMaJ).
We examine CIF as our primary use case, studying how representative benchmarks fail to capture meaningful aspects of model capability.
We complement this analysis with a second ill-defined task, Natural Language to Mermaid Sequence Diagrams (NL2Mermaid), demonstrating how thoughtful benchmark construction and multi-faceted evaluation criteria can reveal actionable insights.
Through these case studies, we identify structural failures in current ill-defined evaluation practices, showing why existing metrics lack interpretability and diagnostic value.
Finally, we distill lessons learned and propose design principles for more robust evaluation of ill-defined tasks.

%% file: latex/ill_defined.tex
\section{Challenges in Evaluating Ill-Defined Tasks}
\label{sec:ill_defined}

We characterize an ill-defined task as a task whose goals, constraints, or evaluation criteria are underspecified, ambiguous, or inherently subjective, such that success cannot be determined by a fixed objective standard and requires interpretation, judgment, or implicit preference.
Leveraging evaluation benchmarks for ill-defined tasks requires selecting data and metrics that account for human interpretation and allow variability.
The evaluation of LLM performance on such tasks is very challenging as they lack fixed correctness conditions, depend on subjective and context-sensitive judgments, and permit multiple valid outputs.

\textbf{Complex Instruction Following (CIF)} is a well-known example of an ill-defined task that requires LLMs to correctly interpret and satisfy multiple, interdependent instructions or constraints such that task success depends on accurate decomposition, ordering, and integration of these instructions.
Examples include multi-constraint text generation, schema-constrained tool calling, and tasks with explicit procedural or ordering requirements.

\textbf{Natural Language to Mermaid Sequence Diagrams (NL2Mermaid)} is an ill-defined task where LLMs generate Mermaid~\citep{mermaid} sequence diagrams from natural language descriptions.
This task is inherently ill-defined because ``correct'' diagramming flows depend on multiple interpretable dimensions: syntactic validity, adherence to syntax conventions, logical consistency of interactions, completeness of specified flows, proper handling of activations, and accurate representation of error and status tracking.
A single natural language input can admit multiple valid diagram structures, this makes it challenging to consistently evaluate model performance.

Despite existing evaluation benchmarks for these tasks, they remain challenging to evaluate reliably.
The fundamental difficulty stems from the inherent ambiguity in ill-defined tasks: multiple valid solutions exist, success criteria are context-dependent, and evaluation metrics must capture subjective judgments.
We summarize the key challenges below:

\begin{itemize}
\item \textbf{Benchmark Limitations and Standardization Gaps.} There is no universally accepted set of benchmarks or evaluation standards for ill-defined tasks, making comparisons across models inconsistent and difficult to interpret.
Many models are post-trained on similar samples as the evaluation test samples, for example, the personas-instruction-following dataset from Tulu3~\citep{lambert2025tulu} follows the same structure as the CIF benchmark IFEval~\citep{zhou2023instructionfollowingevaluationlargelanguage} to obtain models excel on CIF tasks.
However, it may perform poorly on novel, unseen constraint types or complex compositions of constraints, indicating overfitting to specific benchmark formats and limited generalizability.
\item \textbf{Evaluation Metric Challenges.} Traditional metrics (e.g., accuracy, simple compliance) often fail to capture nuanced aspects of ill-defined tasks, as ill-defined tasks do not admit a single correct output or even a finite set of reference answers.
Moreover, task success criteria are often implicit rather than explicit, context-dependent, and interpretable in multiple ways.
Recently researchers rely on LLMaJ to evaluate and score all or partially the LLMs' performance on ill-defined tasks, which bring in new bias depending on judge prompting and model choice.
\item \textbf{Lack of Diagnostic Feedback.}
When a model obtains a low benchmark score, it is often unclear what caused the failure.
For ill-defined tasks, many benchmark metrics aggregate multiple dimensions into a single score, failing to isolate specific reasoning or alignment failures.
This makes it difficult to attribute poor performance to prompt formatting, training data, or training objectives.
The problem is compounded when LLMaJ is used, as judges typically provide scores but rarely offer structured explanations of failures.
Consequently, improving LLM performance on these tasks becomes a matter of chasing higher benchmark scores rather than understanding and fixing underlying capability gaps.
\end{itemize}

Considering the above challenges, the ability of LLMs to perform ill-defined tasks may appear measurable via existing benchmarks, however, existing evaluation approaches are still incomplete, misaligned, or fail to capture the true quality of LLM outputs.

%% file: latex/case_studies.tex
\section{Case Studies in Evaluating Ill-Defined Tasks}
\label{sec:case_studies}

\subsection{Complex Instruction Following}
\label{sec:case_study_cif}

There are many CIF benchmarks for evaluating the instruction following ability of an LLM.
We focus on a few popular benchmarks that are well-recognized by the community, summarized in Table~\ref{tab:instr_benchmarks}.
Although these benchmarks consist of various combinations of instructions and cover a large amount of natural language tasks, they still fail to comprehensively assess the CIF ability of LLMs.
Each benchmark tackles a narrow slice of instruction complexity, and most are not designed to be easily extended to broader scenarios, limiting real-world coverage.
Below, we analyze each benchmark's strengths and limitations, illustrating through concrete examples (Appendix~\ref{sec:appndx_cif}) how existing CIF evaluations fall short:

\begin{table}[h]
\centering
\small
\begin{tabular}{lcccccc}
\toprule
\textbf{Benchmark} 
& \textbf{IFEval}
& \textbf{CB} 
& \textbf{FB} 
& \textbf{MT-Bench} 
& \textbf{HELM} 
& \textbf{StructFlow} \\
\midrule
Instruction complexity      & Low    & High & Med. & High & Med. & High \\
Interaction setting         & S      & S    & S    & M    & S    & M    \\
Evaluation method           & Rule   & Hybrid & Hybrid & LLMaJ & LLMaJ & LLMaJ \\
Objective / reproducible    & Yes    & No   & No   & No   & No   & No \\
Natural instructions        & No     & No   & No   & Yes  & Yes  & Yes \\
Diagnostic feedback         & No     & No   & No   & No   & No   & No \\
\bottomrule
\end{tabular}
\caption{Comparison of complex instruction-following benchmarks.
CB = ComplexBench, FB = FollowBench,
S = single-turn, M = multi-turn, Hybrid = rule-based + LLM-as-a-Judge (LLMaJ).}
\label{tab:instr_benchmarks}
\end{table}

\textbf{IFEval~\citep{zhou2023instructionfollowingevaluationlargelanguage}} consists of a series of formatting instructions and fully tests rule-based instruction following.
Although its metrics are objective and reproducible, the data samples are unnatural, see one example in Appendix~\ref{sec:appndx_cif_ifeval}, and they rarely examine the output content quality, providing limited diagnostic insight beyond pass/fail.
Consider a trivial baseline that ignores semantic content and always outputs a fixed token sequence (e.g., \texttt{AAA}) regardless of instruction semantics.
We observe a score $17.55$ on IFEval for such model, approaching closely to GPT-2 IFEval scores $19.72$~\footnote{\url{https://huggingface.co/datasets/open-llm-leaderboard/gpt2-details}, accessed 2026-01-16. This score was stable across multiple runs. The baseline achieves non-trivial scores primarily on formatting constraints (e.g., length, letter frequency, keyword presence) but not on content-dependent constraints.}.
Since IFEval primarily checks surface-form constraints, degenerate strategies achieve non-trivial scores (e.g., by satisfying a subset of formatting constraints).
This illustrates that IFEval scores can overestimate instruction-following capability when semantic adequacy is not evaluated.

\textbf{ComplexBench (CB)~\citep{wen2024benchmarking}} covers a variety of natural language tasks and composition of instructions, making its data samples diverse and complex.
It also combines rule-based and LLMaJ evaluation methods to evaluate output content quality and instruction compliance via a series of pre-defined scoring questions.
However, the evaluation is significantly limited by the definition of the scoring questions (see example in Appendix~\ref{sec:appndx_cif_cb}), where there may exist bias towards interpretability of the instructions and stochasticity of the LLM judge, demonstrating up to 2\% score variance for the same outputs when the same LLM judge is applied multiple times with different random seeds.
It also suffers language translation issues, the scores vary by 2-4\% if one switches between English and Chinese versions of the data samples.

\textbf{FollowBench (FB)~\citep{jiang-etal-2024-followbench}} exploits the same hybrid evaluation approach, hence, suffers the same issues as CB.
One improvement comparing to the later was a multi-level mechanism to incrementally add a constraint to the initial instruction and provided per-level evaluation, (see example in Appendix~\ref{sec:appndx_cif_fb}).
As expected, nearly all tested LLMs show a decreasing trend as the number of constraints increased.
As with CB, the evaluation provides limited insight into which constraints failed or whether failures stem from instruction interpretation, ordering, or content generation.

\textbf{Other Benchmarks}. MT-Bench~\citep{zheng2023judging} and StructFlowBench~\citep{li2025structflowbench} extend complex instructions to multi-turn user-LLM interactions, but fully rely on LLMaJ to evaluate outputs, which again suffers the same issues as CB and FB.
One improvement they shared with HELM~\citep{liang2022holistic} over other CIF benchmarks is that they curated more natural data samples similar to how human interacts with LLMs.
Other recent CIF benchmarks, such as AgentIF~\citep{qi2025agentif}, FollowRAG~\citep{dong2025toward}, and LIFBench~\citep{wu2024lifbench}, just extend the instruction data coverage to different special natural language tasks, with no contributions to evaluation criteria.

Overall, these attempts show that CIF performance is highly sensitive to prompt structure and surface form.
\cite{wen2024benchmarking} reported that prompt decomposition may adversely affect model ability to follow complex instructions, as evidenced by evaluations on CB.
\cite{dong2025revisiting} shows that IFEval scores can drop by up to 61.8\% when prompts are reformulated (e.g., reordering instructions, rephrasing constraints), suggesting that evaluation sensitivity reflects surface variation rather than stable instruction-following capability.
However, current CIF evaluations provide limited tools to distinguish between failures due to misinterpreted intent, missing knowledge, or compounding errors across constraints, limiting their diagnostic value.


\subsection{Natural Language to Mermaid Sequence Diagrams}
\label{sec:case_study_mermaid}

The NL2Mermaid task exemplifies the challenges of evaluating ill-defined tasks in practice (see Appendix~\ref{sec:appndx_nl2mermaid}).
Initial model outputs frequently exhibit issues such as malformed syntax and missing actor activation or deactivation.
While no widely adopted or standardized benchmark exists for this diagram generation task, recent work has proposed more structured evaluation approaches.

We observed that relying on a single compounded score is insufficient to capture overall model performance.
In particular, \cite{shbita2025mermaidseqbench} proposes evaluating NL2Mermaid outputs along multiple criteria of interest (e.g., \textit{Syntax}, \textit{Logic}, \textit{Completeness}, and \textit{Error and Status Tracking}).
To deepen the investigation and mitigate single-evaluator bias, we further adopted the use of two independent LLMaJ evaluators (DeepSeek-V3~\citep{deepseekai2025deepseekv3technicalreport} and GPT-OSS-120B~\citep{openai2025gptoss120b}) as proposed in that work.
By profiling model outputs across different judges and criteria, we identified recurring weaknesses and intervention opportunities.
One effective intervention was the use of in-context examples, which improved score for certain criteria, though model behavior remained inconsistent and the overall metric did not show improvement.
In this case, using prompt adjustments, we were able to improve the \textit{Logic} score for \textit{Llama~3.1-8B-Instruct}~\citep{grattafiori2024llama3herdmodels} from 67.71 to 80.58 and for \textit{Granite~3.3-8B-Instruct}~\citep{granite2024granite} from 58.90 to 73.90, again, despite a decrease in the overall combined score.
The overall metric combines all criteria with equal weighting and averages judge scores; the apparent decrease in the combined metric despite \textit{Logic} improvement suggests that certain interventions trade off across criteria (e.g., improved logic at the cost of reduced completeness), highlighting the diagnostic value of decomposed evaluation over aggregate scores.
Thus, a tailored solution such as custom prompting or fine-tuning with curated data can be tracked via criteria-specific evaluation rather than an aggregated score.

%% file: latex/discussion.tex
\section{Discussion}
\label{sec:discussion}

Our analysis highlights fundamental limitations in how ill-defined tasks are currently evaluated.
Across both CIF (Section~\ref{sec:case_study_cif}) and NL2Mermaid (Section~\ref{sec:case_study_mermaid}), we find that existing benchmarks and evaluation pipelines often fail to account for the ambiguity, variability, and subjectivity inherent to these tasks, resulting in scores that may fail to measure the intended construct (e.g., true instruction-following ability, diagram generation quality).
As a result, evaluation scores can conflate distinct failure modes, exhibit sensitivity to surface-level prompt variations, and provide limited diagnostic insight into model behavior.

These observations suggest that improving evaluation for ill-defined tasks requires rethinking both benchmark design and metric construction.
Our case studies indicate that metric aggregation (combining multiple criteria into a single score) contributes substantially to poor diagnosis: in NL2Mermaid, decomposed criteria consistently reveal model strengths and weaknesses that aggregate metrics obscure.
However, data coverage and judge variance also matter: IFEval's unnatural instructions and CB's judge variance both limit reliability.
First, evaluation goals must be explicitly defined with clear scoping: benchmarks should document which aspects of model behavior are measured (e.g., constraint compliance, semantic adequacy, structural correctness) and explicitly exclude others, preventing metric conflation.
Second, benchmark data should be designed to reflect realistic variability, including paraphrasing, instruction reordering, and prompt perturbations, to reduce overfitting and expose brittle model behavior.
Third, evaluation metrics should move beyond single aggregate scores toward decomposed analyses that separately capture successful outputs and characteristic failure modes, enabling targeted diagnosis and improvement.
See Appendix~\ref{sec:recommended_suite} for concrete implementation guidance on these principles.

In this context, combining rule-based checks with LLMaJ components can be effective when applied appropriately.
Rule-based evaluation is well suited for decidable properties such as formatting or syntactic validity, while LLMaJ can assess more subjective dimensions when judges are constrained to specific, well-scoped criteria.
A hybrid approach isolates judge noise from genuine model errors: for instance, in NL2Mermaid, syntax validity can be verified deterministically using the Mermaid parser, reserving LLMaJ for semantic and logical criteria, thereby clarifying attribution of failure modes.
However, reliance on LLMaJ without decomposition or reliability analysis risks introducing additional variance and obscuring the sources of model failure.

Overall, our findings indicate that many current evaluations of ill-defined tasks function poorly as diagnostic measurement instruments: they can rank models, but offer limited guidance for understanding or improving model behavior.
As a result, these evaluations are often unstable indicators of underlying capability, underscoring the need for evaluation designs that prioritize robustness, interpretability, and failure-mode analysis.

%% file: latex/appendix.tex
\section{Evaluation Benchmark Examples for Complex Instruction Following}
\label{sec:appndx_cif}

\subsection{An example from IFEval benchmark}
\label{sec:appndx_cif_ifeval}
Listing~\ref{lst:cif_ifeval_prompt} illustrates one sample data from IFEval~\citep{zhou2023instructionfollowingevaluationlargelanguage} benchmark.
We have observed that IFEval scores do not consider the content of the model output, it only checks the requirements.
Here, the scoring framework only checks the frequency of letter ``\texttt{i}'', but not the natural language task for writing a story.
Therefore, the model can pass this test whenever it outputs any single letter.

\begin{lstlisting}[
    label=lst:cif_ifeval_prompt,
    basicstyle=\ttfamily\scriptsize,
    frame=single,
    caption={Example of sample data from IFEval benchmark.},
    captionpos=b,
    numbers=none,
    breaklines=true,
    breakindent=0pt,
    showstringspaces=false,
    linewidth=\columnwidth
]
"key": 201, 
"prompt": "Write a story from a perspective of a man. Include some conversation in the story. Avoid using the letter i more than twice.", 
"instruction_id_list": ["keywords:letter_frequency"], 
"kwargs": [{"let_relation": "less than", "letter": "i", 
"let_frequency": 3}]
\end{lstlisting}


\subsection{An example from ComplexBench benchmark}
\label{sec:appndx_cif_cb}
Listing~\ref{lst:cif_cb_prompt} illustrates one sample data from ComplexBench~\citep{wen2024benchmarking}.
A more natural way to prompt the model would be rewriting the user prompt (inside ``\texttt{instruction\_en}'') into multi-turn conversation.

\begin{lstlisting}[
    label=lst:cif_cb_prompt,
    basicstyle=\ttfamily\scriptsize,
    frame=single,
    caption={Example of sample data from ComplexBench.},
    captionpos=b,
    numbers=none,
    breaklines=true,
    breakindent=0pt,
    showstringspaces=false,
    linewidth=\columnwidth
]
"instruction_en": "- If the user asks about content related to ancient Chinese classics, answer the question according to the user's request, with a character count of no less than 500. Please maintain logical consistency throughout the content of the reply.\n- If the user asks about content related to other types of books, reply with 'I am still learning about this question.'.\n\nUser question: 'Journey to the West', 'All Men Are Brothers', 'the Romance of the Three Kingdoms', and 'the Dream of the Red Chamber' each possess unique artistic techniques in terms of theme, structure, language, characterization, and plot. Please discuss which of these features you appreciate the most and provide a detailed analysis based on the novels' texts.",
"task_types": "Advanced Chinese Understanding",
"constraint_dimensions": [
        "Consistency",
        "Length",
        "Helpfulness",
        "Factuality" ],
"composition_types": [
        "Selection",
        "And"
        ],
"category": "Selection_2"
\end{lstlisting}

To evaluate the model output from the above example, ComplexBench creates a few scoring questions shown in Listing~\ref{lst:cif_cb_score_q}.
Most of the evaluation of these scoring questions involves LLMaJ, other than scoring question 3 (question with ``\texttt{point\_id=3}''), which can be verified by a python script.
One issue with the current scoring system is that the answer of LLMaJ may not be accurate and consistent as some of the scoring question is very subjective, e.g., question 4 and 5 asking about logic consistency and response accuracy.
Another issue with the evaluation system is that if the model outputs ``\texttt{I am still learning about this question}'', it is hard to tell with the current scoring system if the model has knowledge gap but can following instruction or the model just failed to follow instructions.

\begin{lstlisting}[
    label=lst:cif_cb_score_q,
    basicstyle=\ttfamily\scriptsize,
    frame=single,
    caption={Example of socring questions from ComplexBench.},
    captionpos=b,
    numbers=none,
    breaklines=true,
    breakindent=0pt,
    showstringspaces=false,
    linewidth=\columnwidth
]
"scoring_questions": [
{
    "point_id": 0,
    "question_en": "Does the model's response accurately determine that the user's question is related to ancient Chinese classics?",
    "rule": null,
    "constraint_dimensions": [],
    "composition_types": [ "Selection" ],
    "dep": []
}, {
    "point_id": 1,
    "question_en": "Does the model's response discuss which feature of the Four Great Classical Novels it most appreciates?",
    "rule": null,
    "constraint_dimensions": [ "Helpfulness" ],
    "composition_types": [],
    "dep": [ 0 ]
}, {
    "point_id": 2,
    "question_en": "Does the model provide a detailed analysis based on the novels' texts?",
    "rule": null,
    "constraint_dimensions": [ "Helpfulness" ],
    "composition_types": [],
    "dep": [ 0 ]
}, {
    "point_id": 3,
    "question_en": "Does the model's response have a character count of no less than 500 characters?",
    "rule": "length:[500,10000]",
    "constraint_dimensions": [ "Length" ],
    "composition_types": [],
    "dep": [ 0 ]
}, {
    "point_id": 4,
    "question_en": "Does the model maintain logical consistency throughout the content of the reply?",
    "rule": null,
    "constraint_dimensions": [ "Consistency" ],
    "composition_types": [],
    "dep": [ 0 ]
}, {
    "point_id": 5,
    "question_en": "Is the information involved in the content of the model's response accurate and error-free?",
    "rule": null,
    "constraint_dimensions": [ "Factuality" ],
    "composition_types": [],
    "dep": [ 0 ]
}
]
\end{lstlisting}


\subsection{An example from FollowBench benchmark}
\label{sec:appndx_cif_fb}
Listing~\ref{lst:cif_fb_prompt} illustrates one sample data from FollowBench~\cite{jiang-etal-2024-followbench}, where following the initial user instruction (``\texttt{What, according to Milton Friedman, is the role of a business in society?}'') added four other constraints to form a complex user instruction.
Though the instruction is more natural than IFEval and ComplexBench, it lacks complexity on how the instructions are composite as all they are composite sequentially, so the model just needs to execute them one by one.

\begin{lstlisting}[
    label=lst:cif_fb_prompt,
    basicstyle=\ttfamily\scriptsize,
    frame=single,
    caption={Example of sample data from FollowBench.},
    captionpos=b,
    numbers=none,
    breaklines=true,
    breakindent=0pt,
    showstringspaces=false,
    linewidth=\columnwidth
]
"example_id": 2,
"category": "content",
"source": "sharegpt",
"level": 5,
"instruction": "What, according to Milton Friedman, is the role of a business in society? Additionally, analyze its influence on ethical standards in society and identify one possible repercussion on relationships within the community. Please strengthen your argument with one relevant case study and its implications, along with citing one expert opinion or statistical data to support your viewpoint.",
"target": ""
\end{lstlisting}

FollowBench employs an LLMaJ to evaluate the model's output utilizing the evaluation prompt as in Listing~\ref{lst:cif_fb_eval}.

\begin{lstlisting}[
    label=lst:cif_fb_eval,
    basicstyle=\ttfamily\scriptsize,
    frame=single,
    caption={Example of an evaluation prompt from FollowBench.},
    captionpos=b,
    numbers=none,
    breaklines=true,
    breakindent=0pt,
    showstringspaces=false,
    linewidth=\columnwidth
]
Given an initial instruction, we add one content constraint per time and obtain the final instruction with 5 additional constraints.

#Initial Instruction#
What does Milton Friedman believe to be the sole responsibility of business?

#Initial Instruction + 1 constraint#
What is the sole responsibility of business according to Milton Friedman, while also considering its implications on societal ethics?

#Initial Instruction + 2 constraints#
What, according to Milton Friedman, is the single duty of business, while also analyzing its impact on societal ethics and noting one potential consequence on community relations?

#Initial Instruction + 3 constraints#
What, in the view of Milton Friedman, constitutes the primary responsibility of a business? Additionally, evaluate its influence on societal ethics and highlight one possible repercussion on community relations, incorporating one relevant case study for context.

#Initial Instruction + 4 constraints#
What, according to Milton Friedman, is the main duty of a business? Furthermore, assess its impact on societal ethics and pinpoint one potential consequence on community relations, ensuring to substantiate your answer with one pertinent case study and its implications.

#Initial Instruction + 5 constraints#
What, according to Milton Friedman, is the role of a business in society? Additionally, analyze its influence on ethical standards in society and identify one possible repercussion on relationships within the community. Please strengthen your argument with one relevant case study and its implications, along with citing one expert opinion or statistical data to support your viewpoint.

#Answer of Initial Instruction + 5 constraints#
<MODEL'S_OUTPUT>

#System#
1) Please identify all 5 added constraints.
2) For the 5 added constraints, discriminate if the #Answer of Initial Instruction + 5 constraints# satisfies each constraint.
3) In the final line, only output a Python LIST with 5 elements ('YES' or 'NO') indicating whether the answer satisfies each constraint.
\end{lstlisting}


\section{Natural Language to Mermaid Sequence Diagram Example}
\label{sec:appndx_nl2mermaid}

Listing~\ref{lst:uml-sequence-nl} illustrates one representative NL2Mermaid test case showing the natural language input specification for that task.
The listing shows the full input, including the purpose of the diagram, the main components, and the interactions in the form of an ordered sequence.

\begin{lstlisting}[
    label=lst:uml-sequence-nl,
    basicstyle=\ttfamily\scriptsize,
    frame=single,
    caption={Natural language specification for ``Uploading Documents with Secure Storage'' task.},
    captionpos=b,
    numbers=none,
    breaklines=true,
    breakindent=0pt,
    showstringspaces=false
]
Purpose: Uploading Documents with Secure Storage
Main Components: User, Mobile App, BFF, Azure AD
Interactions:
1. User Action: User uploads a document through the mobile app.
2. Mobile App: Sends the document along with the session token to the BFF.
3. The BFF validates the token with Azure AD.
4. On successful upload, the BFF returns a confirmation to the app. If the user is unauthorized or the file exceeds size limits, an appropriate error is returned.
\end{lstlisting}

Listing~\ref{lst:uml-sequence-mermaid} shows the corresponding erroneous Mermaid syntax generated by the LLM.
Here, the output diagram exhibited several issues that were recurrent across the dataset: inconsistent activation and deactivation markers, malformed or misplaced error-handling logic, and incomplete responses.

\begin{lstlisting}[
    label=lst:uml-sequence-mermaid,
    basicstyle=\ttfamily\scriptsize,
    frame=single,
    caption={Mermaid syntax illustrating a faulty ``Uploading Documents with Secure Storage'' flow, showing activation inconsistencies, incorrect error-handling, and incomplete actor responses.},
    captionpos=b,
    numbers=none,
    breaklines=true,
    breakindent=0pt,
    showstringspaces=false
]
sequenceDiagram
title Uploading Documents with Secure Storage

participant User
participant App as Mobile App
participant BFF
participant AAD as Azure AD

User->>App: Select document & upload
activate App
App->>BFF: POST /upload (file, token)
activate BFF
BFF->>AAD: validate (token)
activate AAD
AAD-->>BFF: 200 OK
deactivate App

alt success
  BFF-->>Application: 201 Created
  App-->>User: Done
  deactivate BFF
else unauthorized
  AAD-->>BFF: 401 Unauthorized
  BFF-->>App: 401
else too large
  BFF-->>App: 413 Payload Too Large
end
\end{lstlisting}


\section{Recommended Diagnostic Evaluation Suite}
\label{sec:recommended_suite}

To address the challenges identified in this paper, we recommend that benchmark designers implement the following minimal diagnostic suite when evaluating ill-defined tasks:

\begin{enumerate}
  \item \textbf{For Complex Instruction Following:} Create explicit rule-based checkers for each constraint type (formatting, length, keyword presence, content constraints) and document which constraints are evaluated by deterministic rules versus LLMaJ. This separation clarifies which failures stem from factual/semantic gaps versus constraint misinterpretation.
  \item \textbf{For structured generation tasks (e.g., NL2Mermaid):} Implement separate rule-based syntax validators (e.g., grammar parsers, schema validators) before invoking LLMaJ for semantic or logical criteria. This hybrid approach isolates judge noise from genuine structural errors.
  \item \textbf{Judge calibration and transparency:} Report inter-judge agreement metrics, document the exact prompt template used for LLMaJ evaluation, specify judge model versions and hyperparameters (temperature, seed), and compute confidence intervals on final scores. This enables future researchers to reproduce evaluations and identify variance sources.
\end{enumerate}

These minimal elements enable benchmark reproducibility, facilitate failure attribution, and can provide the foundation for principled benchmark iteration.
